\documentclass{article} 
\usepackage{iclr2027_conference,times}

\usepackage{amsmath,amsfonts,bm}

\def\eqref#1{equation~\ref{#1}}

\def\1{\bm{1}}

\DeclareMathAlphabet{\mathsfit}{\encodingdefault}{\sfdefault}{m}{sl}
\SetMathAlphabet{\mathsfit}{bold}{\encodingdefault}{\sfdefault}{bx}{n}

\usepackage{hyperref}
\usepackage{url}
\usepackage{graphicx}
\usepackage{xspace}
\usepackage{amssymb,amsthm}
\usepackage{xcolor}
\usepackage{algorithm}
\usepackage{algorithmic}
\usepackage{enumitem}
\usepackage{xspace}
\usepackage{tcolorbox}
\tcbuselibrary{skins,breakable}

\definecolor{evoblue}{HTML}{356B9A}
\definecolor{evobluebg}{HTML}{F2F7FB}
\newtcolorbox{promptbox}[1]{enhanced, breakable, colback=evobluebg, colframe=evoblue, colbacktitle=evoblue, coltitle=white, fonttitle=\bfseries\small, title={#1}, arc=1.5mm, boxrule=0.6pt, left=6pt, right=6pt, top=4pt, bottom=4pt, toptitle=2pt, bottomtitle=2pt, fontupper=\small\raggedright, before skip=4pt, after skip=6pt}

\usepackage{wrapfig}
\usepackage{booktabs}
\usepackage{pifont}

\makeatletter
\newcommand{\captionof}[1]{\def\@captype{#1}\caption}
\makeatother

\newcommand{\cmark}{\ding{51}}
\newcommand{\xmark}{\ding{55}}
\graphicspath{{figures/}}

\newcommand{\agent}{HypAgent\xspace}
\newcommand{\benchmark}{OpenHypBench\xspace}
\newcommand{\task}{Open-ended Hypothesis Discovery\xspace}

\newtheorem{definition}{Definition}

\newif\ifshowcomments
\showcommentstrue 

\ifshowcomments
    \newcommand{\nc}[1]{{\color{blue}\bfseries [NC: #1]}}
    
\else
    \newcommand{\nc}[1]{}
\fi

\title{\benchmark: Benchmarking \task}

\title{Searching for What Matters: Open-Ended Hypothesis Discovery Through Agentic Data Exploration}
\title{Open-Ended Hypothesis Discovery Through Agentic Data Exploration}
\title{Open-Ended Hypothesis Discovery \\ via Agentic Search}

\title{Searching for What Matters:\\ Open-Ended Discovery of Multiple\\ \mbox{Hypotheses} with Agentic Data Exploration}

\author{%
  \makebox[\dimexpr\textwidth-2\tabcolsep\relax][c]{%
    \bfseries
    Sanchit Kabra$^{1}$\thanks{Work done during an internship at Amazon.}\hspace{1.5em}%
    Nurendra Choudhary$^{2}$\hspace{1.5em}%
    Jiangshu Du$^{2}$\hspace{1.5em}%
    Eddie Huang$^{2}$%
  }\\[0.25em]
  \makebox[\dimexpr\textwidth-2\tabcolsep\relax][c]{%
    \bfseries
    Danai Koutra$^{2}$\hspace{1.5em}%
    Chandan K.\ Reddy$^{1,2}$%
  }\\[0.75em]
  \makebox[\dimexpr\textwidth-2\tabcolsep\relax][c]{%
    \large
    $^{1}$Virginia Tech \qquad $^{2}$Amazon
  }%
}

\iclrfinalcopy 
\begin{document}

\maketitle

\begin{abstract}
Discovering what drives an outcome from data is central to science and often requires deciding what is worth investigating before a specific hypothesis is known. Yet most existing hypothesis-generation settings substantially narrow this search by providing a research question, target relationship, or curated context. We introduce \textbf{\task}, where a system receives only a large tabular dataset and an outcome and must discover multiple data-supported hypotheses without knowing in advance the relevant variables, functional forms, conditions, or number of underlying mechanisms. To study this problem, we introduce \textbf{\benchmark}, a benchmark built from structural causal models grounded in real-world data schemas, containing 351 planted hypotheses across 15 tasks and five domains, with 53--65 variables and 1M--10M rows per task. The causal model is hidden from the discovery method and retained only by the evaluator, allowing proposed hypotheses to be tested through interventions and separating observational discovery from causal validation. We further propose \textbf{\agent}, which treats the dataset as an interactive environment and performs iterative search by proposing candidate directions, triaging and refining promising hypotheses, and removing explained signal so that each accepted discovery redirects subsequent search. \agent achieves 0.47 precision, 0.51 recall, and 0.49 F1, compared with 0.35 F1 for the strongest baseline. Our analysis shows that many failures occur after a relevant variable has already been surfaced, suggesting that correctly characterizing a mechanism's functional form and conditions, rather than merely identifying relevant variables, remains a key challenge in open-ended discovery.
\end{abstract}

\section{Introduction}

Data-driven discovery often begins before there is a hypothesis to test: one must first decide \emph{what is worth investigating}. This remains difficult even as large language models and agents increasingly automate downstream work such as coding, data analysis, and hypothesis testing
\citep{yang2024sweagent,guo2024dsagent,jing2025dsbench,chen2025scienceagentbench}.
Consider a large tabular e-commerce dataset. Purchase behavior may simultaneously depend on a discount threshold, delivery-time sensitivity, and a marketing effect that appears only for a particular customer segment. A scientist may not know beforehand that any of these mechanisms exist, which variables define them, or how many distinct mechanisms are present.

Most existing hypothesis-generation settings substantially narrow this search in advance by providing a research question, discovery goal, prediction target, treatment, or curated scientific context
\citep{majumder2025discoverybench,yang2025moosechem,liu2026researchbench,zhou2024hypothesis,qi2024biomedical,liu2025literature}.
We instead study \textbf{\task}: given only a tabular dataset $D$ and an outcome $Y$, discover a set
$H=\{h_1,\ldots,h_K\}$ of data-supported hypotheses without guidance about which variables to investigate, what functional relationships to consider, under what conditions they hold, or even how many distinct hypotheses $K$ are present.
Here, \emph{open-ended} refers to the discovery target being unspecified while the input itself remains structured tabular data.

\begin{figure}[t]
    \centering
    \includegraphics[width=\textwidth]{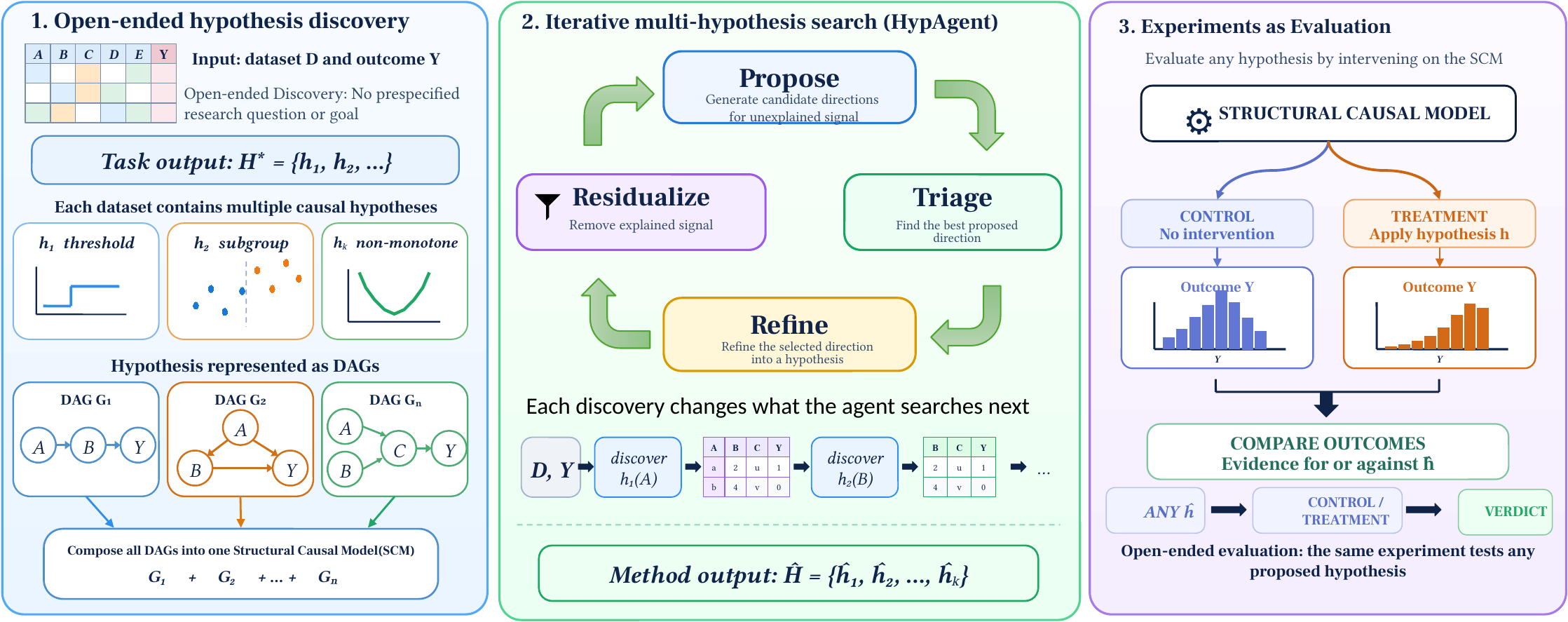}
    \vspace{-0.5cm}
    \caption{
\textbf{Overview of Open-Ended Hypothesis Discovery.}
\textbf{Left:} Given only a tabular dataset $D$ and outcome $Y$, the goal is to discover multiple unknown hypotheses.
\textbf{Middle:} \agent{} iteratively proposes, triages, refines, and residualizes hypotheses so that each discovery guides the next search.
\textbf{Right:} \benchmark{} evaluates proposed hypotheses through interventions on the hidden structural causal model.
}
    \label{fig:overview}
\end{figure}

This setting creates a difficult search problem. The space of possible hypotheses grows rapidly with the number of variables, nonlinear relationships, interactions, thresholds, and subgroup conditions. Moreover, multiple mechanisms may coexist in the same data, so discovering one strong relationship does not imply that the discovery process is complete. A system must therefore do more than generate a plausible hypothesis once. Existing method families address only parts of this problem. Statistical and causal-discovery methods can operate over large datasets, but typically require the treatment or hypothesis family to be specified in advance, or return structural relationships rather than complete hypotheses
\citep{agrawal1993mining,chernozhukov2018double,wager2018estimation,Spirtes1993CausationPA,Chickering2002OptimalSI}.
LLM-based hypothesis generators are more expressive, but generally reason over examples or summaries placed in the model context
\citep{zhou2024hypothesis,yang2025moosechem,liu2025literature}.
Data-science agents can execute code on large datasets, but are typically given an external objective, such as optimizing a predictive metric
\citep{guo2024dsagent,hong-etal-2025-data}.
Open-ended discovery instead requires the system itself to decide what is worth investigating next.

Evaluation poses a second challenge. Reference-set matching alone is insufficient because hypotheses involving similar variables may differ in functional form, direction, conditions, or effect magnitude, while a valid proposed hypothesis need not be phrased identically to a planted reference. To make this problem measurable, we introduce \textbf{\benchmark}, a benchmark built from structural causal models grounded in schemas and empirical variable distributions of real-world datasets. Each task contains multiple heterogeneous mechanisms composed into a structural causal model (SCM), yielding large datasets in which different effects can overlap, interact, or apply only under particular conditions. Across 15 tasks that span five domains, \benchmark{} contains 351 planted hypotheses, 53-65 variables per task, and datasets ranging from 1M to 10M rows.
The underlying SCM is hidden from the discovery method and retained only by the evaluator. For a given proposed hypothesis, the evaluator performs the corresponding intervention on its variables and measures the resulting change in the outcome. Consequently, the discovery method uses observational data to generate and support candidate hypotheses, while \benchmark{} independently tests their interventional validity
\citep{pearl2009causality,pearl2019seventools}.

To solve the \task task, we introduce \textbf{\agent}, an agent designed to iteratively search for multiple hypotheses in datasets too large to place directly in an LLM context. Rather than treating the dataset as text, \agent{} treats it as an environment that can be queried programmatically. Each discovery cycle first proposes candidate directions, uses lightweight analyses over sampled observations to triage promising candidates, and then queries the full dataset to refine the selected direction into a more specific data-supported hypothesis. Accepted hypotheses are used to model and remove explained outcome variation before the next discovery cycle. The central principle is that \emph{each accepted discovery changes what is searched next}. Inspired by matching pursuit
\citep{mallat1993matching}, this residual-driven procedure progressively redirects the search to the unexplained signal, as illustrated in Figure~\ref{fig:overview}.

Across \benchmark{}, \agent{} achieves an $F_1$ of 0.49, compared to 0.35 for the strongest baseline. Its recall advantage persists under matched hypothesis-proposal budgets. Our analysis further shows that identifying a relevant variable is often insufficient because the majority of failures occur after a promising variable has already surfaced. These results suggest that open-ended hypothesis discovery requires both an efficient search over candidate mechanisms and careful refinement of the hypotheses that the search uncovers, while overall performance leaves substantial room for improvement. The contributions of this work are:
\begin{itemize}[leftmargin=*]
\vspace{-.8em}
\setlength\itemsep{-0.2em}

    \item \textbf{New Task:} We formulate \textbf{\task}, where a system must discover multiple hypotheses from a large tabular dataset and outcome without being given the variables, functional forms, conditions or number of mechanisms to investigate.

    \item \textbf{New Benchmark:} We introduce \textbf{\benchmark}, comprising 351 planted hypotheses across 15 tasks and five domains, together with an interventional evaluation framework that tests the proposed hypotheses in addition to measuring recovery against the planted reference set.

    \item \textbf{Agentic Framework:} We propose \textbf{\agent}, an iterative residual-driven agent that treats the dataset as an interactive environment and uses each accepted discovery to redirect the subsequent search towards unexplained outcome variation. Our experiments further identify mechanism characterization, rather than merely surfacing relevant variables, as a key remaining challenge.
\end{itemize}

\vspace{-0.3cm}
\section{Related Work}
\vspace{-0.3cm}
\textbf{Benchmarks for hypothesis discovery.} Automated scientific discovery has historically been studied in constrained settings where the hypothesis space is defined in advance. Early systems recovered physical laws or causal structures from data but operated over fixed variable sets and predefined hypothesis families \citep{doi:10.1126/science.1165620,doi:10.1126/science.1165893}, and causal discovery benchmarks inherited this structure by evaluating recovery of a target graph over a known variable set \citep{sachs2005,scutari2010bnlearn}. More recent benchmarks broaden the range of expressible hypotheses but still evaluate against predefined targets: findings drawn from published scientific work \citep{majumder2025discoverybench,liu2025hypobench}, expert-authored questions or known solutions \citep{gu-etal-2024-blade,chen2025scienceagentbench,jing2025dsbench,jansen2024discoveryworld}, or hypotheses generated from provided context \citep{qi2024biomedical,liu2026researchbench}. \citet{chen2026hypospace} evaluate hypothesis sets within finite and exactly enumerable spaces. These designs do not directly evaluate an arbitrary supported proposal that falls outside the predefined target set, which is required in our open-ended setting. Table \ref{tab:benchmark_positioning} summarizes previous hypothesis discovery benchmarks in the context of open-ended hypothesis discovery.
\vspace{-0.3cm}
\begin{table}[h]
\centering
\vspace{-0.4cm}
\caption{Comparison of existing hypothesis discovery benchmarks. \emph{Scale} denotes the maximum number of rows in a benchmark dataset. \emph{New targets} indicates that the evaluated data and target effects are newly generated rather than published targets.}
\label{tab:benchmark_positioning}
\small
\setlength{\tabcolsep}{4.5pt}
\renewcommand{\arraystretch}{1.08}
\resizebox{\linewidth}{!}{%
\begin{tabular}{@{}lccccc@{}}
\toprule
\textbf{Benchmark} &
\textbf{Scale} &
\textbf{Discovery setting} &
\shortstack{\textbf{Open-ended}\\\textbf{evaluation}} &
    \shortstack{\textbf{New}\\\textbf{targets}} &
\textbf{Causal} \\
\midrule
Context-based generation
\citep{qi2024biomedical}
& 3K
& Curated context
& \xmark
& \xmark
& \xmark \\

Data-driven analysis
\citep{gu-etal-2024-blade}
& $\sim$100K
& Prespecified question
& \xmark
& \xmark
& \xmark \\

Hypothesis suites
\citep{liu2025hypobench,majumder2025discoverybench}
& $\sim$100K
& Prespecified goal
& \xmark
& \xmark
& \xmark \\

Causal suites
\citep{sachs2005,scutari2010bnlearn}
& 10K
& Prespecified goal
& \xmark
& \xmark
& \cmark \\

\midrule
\textbf{\benchmark{} (ours)}
& \textbf{10M}
& \textbf{Open-ended}
& \cmark
& \cmark
& \cmark \\
\bottomrule
\end{tabular}
}
\end{table}

\textbf{Methods for hypothesis generation.}
\label{sec:methods-related}
Prior methods divide along two axes, whether they generate hypotheses in open-ended settings and whether they \emph{scale without placing data in context}. LLM-based hypothesis generators~\citep{zhou2024hypothesis,liu2025literature} typically receive a research question or scientific context and reason over examples or summaries placed in the prompt. LLM data-science agents such as DS-Agent~\citep{guo2024dsagent} and Data Interpreter~\citep{hong-etal-2025-data} can execute code over external datasets, but they optimize a fixed, externally specified objective rather than deciding what is worth investigating. Statistical pipelines rank associations or predictive features, while causal-structure methods such as PC and GES estimate graphical relationships over a specified variable set~\citep{Spirtes1993CausationPA,Chickering2002OptimalSI}. Neither directly returns the complete hypotheses required here, including their conditions, directions, and magnitudes.


\vspace{-0.3cm}
\section{Problem Formulation}
\label{sec:problem}
\vspace{-0.3cm}
Let $D=\{(\mathbf{x}_i,y_i)\}_{i=1}^{n}$ denote $n$ observations over
features $X=\{X_1,\ldots,X_p\}$ and a designated outcome $Y$. A hypothesis $h$ is a testable claim about how intervening on one or more
variables affects the outcome under specified conditions. We represent it as
$
h=(X_h,\psi,d,c,m_h),
$
where $X_h\subseteq X$ contains the variables that participate in the mechanism,
$T_h\subseteq X_h$ denotes the variables intervened on to test the claim,
$\psi$ specifies their structural relationship with the outcome, such as a
monotone, thresholded, context-dependent or mediated relationship,
$d\in\{+,-,\mathrm{mixed}\}$ denotes the claimed direction of the effect, with
$\mathrm{mixed}$ indicating that the direction changes across the regions or
conditions specified by $\psi$ and $c$. The term $c$ specifies the condition
or population for which the claim is valid and $m_h\geq 0$ denotes
the claimed absolute change in the probability of the outcome. For
example,
\[
h_1=
\left(
\{\texttt{recency}\},
\texttt{threshold}(\texttt{recency}>47),
+,
\text{all customers},
0.06
\right)
\]
states that, for all customers, purchase recency above
47 days increases the probability of churn by six percentage points.
Together, the components of $h$ distinguish hypotheses that involve similar
variables but make different claims about their conditions,
direction, or magnitude.

\begin{definition}[\task]
Given an observational dataset $D$ and a designated outcome $Y$, the goal of
\task is to produce a set of testable causal hypotheses
$
\widehat H
=
\mathcal{A}(D,Y)
=
\{\hat h_1,\ldots,\hat h_{\widehat K}\}
$
about distinct mechanisms affecting $Y$. The method is not given the number
of mechanisms, intervention variables, participating variables, structural
relationships, conditions, directions, or effect magnitudes to search for.
\end{definition}
For each task, $H^\star=\{h_1^\star,\ldots,h_{K^\star}^\star\}$ denotes the set
of $K^\star$ planted mechanisms hidden from the discovery method.
Because \task seeks multiple hypotheses rather than a single answer, we
distinguish \emph{mechanism identity} from \emph{hypothesis correctness}. Two
hypotheses correspond to distinct mechanisms when they differ in their
variable set $X_h$ or structural relationship $\psi$, or when their conditions $c$ are not equivalent. Differences in wording, direction, or effect magnitude alone do not define a new mechanism. Instead, direction and magnitude determine whether a claim about that mechanism is correct. Section~\ref{sec:benchmark} describes how \benchmark instantiates this task
using observational datasets with multiple hidden causal mechanisms and an
interventional evaluator.

\vspace{-0.3cm}
\section{\benchmark}
\label{sec:benchmark}
\vspace{-0.3cm}
Evaluating \task requires more than matching generated hypotheses against a fixed answer set. A proposal may make a valid claim that is not identical to a planted hypothesis, while a proposal involving the correct variables may still be wrong in its conditions, direction, or magnitude. Moreover, real observational datasets do not provide interventional ground truth for arbitrary proposed claims. We therefore construct \benchmark as a collection of controlled data-generating worlds in which the underlying causal mechanisms are known to the evaluator but hidden from the discovery method.
\vspace{-0.3cm}
\subsection{Constructing Multi-Mechanism Data-Generating Worlds}
\label{sec:benchmark_construction}
\begin{figure}[t]
    \centering
    \begin{minipage}[t]{0.49\textwidth}
        \vspace{0pt}
        \centering
        {\scriptsize
        \setlength{\tabcolsep}{2.5pt}
        \renewcommand{\arraystretch}{1.08}
        \begin{tabular}{@{}llcc@{}}
            \toprule
            Domain & Source schemas & Feat. & Mech. \\
            \midrule
            Census & ACS, Adult, CPS & 63--64 & 73 \\
            E-commerce & \begin{tabular}[t]{@{}l@{}}H\&M, Instacart,\\Online Retail\end{tabular} & 61--63 & 68 \\
            Education & \begin{tabular}[t]{@{}l@{}}OULAD, Student\\Performance, xAPI\end{tabular} & 58--65 & 70 \\
            Finance & \begin{tabular}[t]{@{}l@{}}Give Me Some Credit,\\LendingClub, Taiwan\end{tabular} & 58--65 & 70 \\
            Health & \begin{tabular}[t]{@{}l@{}}BRFSS, Diabetes130,\\NHANES\end{tabular} & 53--65 & 70 \\
            \midrule
            Total & 15 tasks & 53--65 & 351 \\
            \bottomrule
        \end{tabular}}
        \captionof{table}{Summary of \benchmark. Each domain contains 3 tasks with 1M--10M rows. Dataset citations and licenses are reported in Appendix~\ref{app:benchmark_details}.}
        \label{tab:benchmark_summary}
    \end{minipage}
    \hfill
    \begin{minipage}[t]{0.48\textwidth}
        \vspace{0pt}
        \centering
        \includegraphics[width=\linewidth]{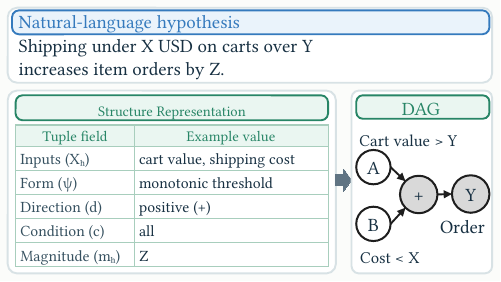}
        \vspace{-0.5cm}
        \captionof{figure}{Example mapping from a natural-language hypothesis to the structured representation used by \benchmark{} and its corresponding mechanism DAG.}
        \label{fig:hypothesis-to-dag}
        
    \end{minipage}
    \vspace{-0.3cm}
\end{figure}

\vspace{-0.2cm}
Each \benchmark task begins from a public tabular dataset. We retain its variable vocabulary, data types, semantics, and empirical marginal distributions, but construct new causal mechanisms and generate new observations specifically for the benchmark. This preserves realistic variable semantics and marginal ranges while allowing the causal structure and interventional ground truth to be controlled. \benchmark contains 15 tasks spanning census, e-commerce, education, finance, and health, with three distinct source schemas per domain. Tasks contain 53--65 visible features and 1M--10M generated observations, and the benchmark contains 351 planted mechanisms in total. Table~\ref{tab:benchmark_summary} summarizes the domains and source schemas; Appendix~\ref{app:benchmark_details} provides the complete task inventory, including outcomes, licenses, scale, feature counts, and individual mechanism counts.

To instantiate heterogeneous discovery challenges, \benchmark uses seven hypothesis families: monotone threshold, non-monotone, derived-feature, interaction-reversal, subpopulation-specific, Simpson's-reversal, and mediation. These families define the mechanisms used to construct the SCM, and discovery methods are not required to predict a family label. Appendix~\ref{app:functional_forms} defines each family and reports its frequency. Candidate mechanisms are proposed using Claude Opus 4.8 with the prompt reported in Appendix~\ref{app:benchmark_construction_prompt}. Each retained hypothesis is represented using the formulation in Section~\ref{sec:problem} and instantiated as an individual causal DAG, as illustrated in Figure~\ref{fig:hypothesis-to-dag}. The individual DAGs are then composed into a single task-level structural causal model (SCM). Specifications containing directed cycles, undefined parents, missing moderators, or incompatible operations on variable types are rejected before data generation.

Variables in the resulting SCM serve different roles, including causes, confounders, moderators, mediators, background variables, decoys, and null variables. In particular, decoys may correlate with causal variables through shared upstream structure while having no direct outcome mechanism, creating plausible but non-causal directions for discovery methods to distinguish from true mechanisms. The SCM jointly generates all observations in a task. Root variables are sampled from the empirical distributions of the corresponding source variables, and remaining variables are generated in topological order according to structural assignments
$
V_j := f_j\!\left(\mathrm{Pa}(V_j),\epsilon_j\right),
$
where $\mathrm{Pa}(V_j)$ denotes the parents of $V_j$ and $\epsilon_j$ captures exogenous variation. A mechanism may be active only under particular conditions or within specific regions of the feature space. Consequently, multiple heterogeneous mechanisms may coexist in the same dataset while all observations arise from one shared SCM. Importantly, the planted mechanisms participate directly in the data-generating process rather than being attached to generated observations as post hoc annotations.
\vspace{-0.3cm}
\subsection{Interventional Evaluation of Proposed Hypotheses}
\label{sec:interventional_evaluation}

The SCM used to generate each task is hidden from every discovery method and retained only by the \benchmark evaluator. Methods produce hypotheses in the common representation defined in Section~\ref{sec:problem} or their outputs are parsed into that representation. For each proposed hypothesis, the evaluator constructs a deterministic experiment plan specifying the intervention variables, values, and any conditions needed to test the stated claim. Appendix~\ref{app:evaluation_protocol} describes proposal normalization, deduplication, and recall matching.

Let $T_h$ denote the intervention variables of hypothesis $h$. Executing $do(T_h=t)$ replaces the structural assignments of those variables with the intervention values $t$, removes their incoming edges, and regenerates their descendants in topological order while leaving the remainder of the SCM unchanged~\citep{pearl2019seventools}. Each hypothesis is evaluated using 60,000 paired samples under the required intervention conditions. For a given hypothesis, the resulting causal contrast can be written as
$
\tau_h =
\mathbb{E}\!\left[Y \mid do(T_h=t_1),c\right]
-
\mathbb{E}\!\left[Y \mid do(T_h=t_0),c\right],
$
where $c$ denotes any population restriction specified by the hypothesis. Here, $\tau_h$ is the signed causal contrast measured by the evaluator, whereas $m_h$ is the nonnegative magnitude claimed in the hypothesis representation. Conditions used for such comparisons must involve variables that are not descendants of $T_h$. Experiment plans conditioning on descendants of an intervened variable are rejected. Other structural forms induce corresponding form-specific experiments and contrasts, such as evaluating interaction reversals across moderator values or non-monotone effects across multiple treatment regions. The resulting interventional response is compared with the structural form, direction, condition, and magnitude of the claim. Thus, the discovery method reasons only from observational data, while causal support is assessed separately through interventions on the hidden SCM.

\textbf{Evaluation metrics.} We evaluate methods using \emph{precision}, \emph{recall}, and $F_1$ score. These quantities capture complementary aspects of open-ended discovery: precision measures whether the distinct hypotheses proposed by a method are supported by intervention, while recall is the fraction of the $K^\star$ planted mechanisms recovered by the method. Before computing precision, proposals are deduplicated according to the mechanism-identity rule in Section~\ref{sec:problem}, retaining the first proposal encountered for each identity without using interventional support to select the representative. A supported proposal need not match a planted hypothesis and may therefore receive precision credit even if it lies outside the planted reference set. Recall measures recovery of the set of planted mechanisms $H^\star$. The supported proposals are first shortlisted against the planted mechanisms using their intervention variables, interventional response structure, conditions, and magnitude of the effect. Claude Opus 4.8 then performs the final mechanism-equivalence judgment. Multiple proposals matched to the same planted mechanism count only once. Unless otherwise stated, a claimed effect magnitude $m_h$ is considered correct when it differs by at most 10\% from the corresponding interventional effect magnitude. Refer to Appendix~\ref{app:evaluation_protocol} for further details on implementation.

\vspace{-0.3cm}
\section{\agent: Residual-Driven Hypothesis Search}
\label{sec:hypagent}
\vspace{-0.3cm}
In practical settings, \task requires reasoning over datasets far larger than an LLM's context window. Directly ingesting such data is infeasible, while subsampling, chunking, or reducing the data to aggregate statistics can disturb the distributional structure and miss dependencies important for discovery. \agent therefore uses sampled observations to propose and triage candidate directions, while residualization and refinement operate over the full dataset. The LLM queries the data through code and receives only the resulting statistics or summaries rather than the complete table in its context.

A key challenge in generating hypotheses in an open-ended setting is that strong effects can repeatedly dominate the search, causing a method to rediscover variations of hypotheses it has already identified, while weaker mechanisms remain unexplored. We therefore design \agent around a simple principle: \emph{each accepted discovery should change what is searched next}. We operationalize the search using a matching-pursuit-inspired procedure~\citep{mallat1993matching}. Matching pursuit decomposes a signal by repeatedly identifying a component that explains part of it, removing its contribution, and continuing on the residual. In our setting, the signal is the outcome $Y$, and the components are candidate hypotheses. Each round consists of three stages: \emph{residualization} removes the effects already discovered, \emph{triage} identifies which proposed candidate best explains the remaining signal, and \emph{refinement} recovers conditional structure and validates the resulting hypothesis. The agent operates only on observational data and does not access the hidden SCM or the interventional evaluator used by \benchmark. Refer to Algorithm~\ref{alg:hypagent} for pseudocode of \agent.

\textbf{Residualization.} To address the challenge of recovering hypotheses beyond those that dominate the search, during each iteration \agent removes the variation explained by accepted hypotheses, thereby changing the search target after each discovery. Let $S_k$ denote the hypotheses accepted after round $k$. Each hypothesis $h\in S_k$ contributes a representation $\phi_h(X)$ of its named variables, and $\Phi(S_k)$ denotes their combined representation. Appendix~\ref{app:residualization} specifies the variable encodings. Using the class $\mathcal{F}_{\mathrm{lin}}$ of linear predictors, we define
$
    \hat f_k=\arg\min_{f\in\mathcal{F}_{\mathrm{lin}}}
    \left\|Y-f\!\left(\Phi(S_k)\right)\right\|_2^2,
    \qquad
    \widetilde Y_k=Y-\hat f_k\!\left(\Phi(S_k)\right).
$
The next iteration operates on $\widetilde Y_k$ rather than $Y$. Consequently, hypotheses already captured by $S_k$ contribute less signal to subsequent search, shifting attention toward mechanisms that explain previously uncovered variation.
Residualization removes fitted observational variation rather than a known causal mechanism. It therefore redirects search toward signal not yet represented by the accepted hypotheses, without assuming a unique causal decomposition.

\textbf{Proposal \& Triage.}
The space of candidate hypotheses that explain the residual is too large to traverse systematically. \agent leverages the world knowledge of LLMs to propose a set of candidate hypotheses using the dataset and the hypotheses accepted so far. Rather than spending resources on each candidate hypothesis, \agent first performs a lightweight data-driven triage by querying the data to determine which candidate hypothesis explains the largest fraction of the remaining outcome variation. This step provides exploration in the discovery process by asking which broad direction explains most of the residual variation.

For each candidate variable $T$, we compute its empirical squared correlation ratio with the current residual,
$
s(T)
=
\frac{
    \mathrm{Var}\!\left(
        \mathbb{E}[\widetilde Y_k \mid \mathrm{bin}(T)]
    \right)
}{
    \mathrm{Var}(\widetilde Y_k)
},
$
where $\mathrm{bin}(T)$ denotes quantile bins for continuous variables and observed categories for categorical variables. \agent uses a tunable threshold \(\epsilon\), requiring \(s(T)\geq\epsilon\) for a candidate to be retained for further analysis. Among candidates that clear this threshold, the highest-scoring candidate is selected. Triage therefore allows \agent to cheaply filter a broad set of directions before committing to analyze a single candidate.

\textbf{Refinement.} 
Triage identifies a promising starting point, but the selected direction is based on a limited triage sample and remains underspecified. Committing to it directly would therefore favor a simple explanation when the underlying relationship may depend on conditions, subpopulations, or interactions. Once a candidate $T^\star$ is selected, \agent shifts from broad search to exploitation where it queries the data to refine $T^\star$ towards a hypothesis that most accurately explains the remaining outcome. These queries inspect the candidate across different regions of its distribution, subpopulations, and potential conditioning variables, progressively refining the initial direction into a more specific hypothesis. Refinement seeks to answer: \textit{Given a direction, what more specific formulation explains more of that variation in the outcome residual?}

Once \agent narrows down to a specific hypothesis, two sources of ambiguity still remain. First, the relationship associated with $T^\star$ may change depending on another variable. To test this, \agent leverages the world knowledge and reasoning capabilities of LLMs to identify candidate moderators $M$ and fits
$\widetilde Y_k = \beta_0 + \beta_T T^\star + \beta_M M + \beta_{TM}(T^\star \times M)$.
The interaction coefficient $\beta_{TM}$ measures whether the relationship associated with $T^\star$ changes with $M$. This helps detect context-dependent associations that can appear weak or disappear when viewed only in aggregate. A second ambiguity is confounders, where a candidate can explain residual variation because it is associated with another variable that is responsible for the signal. \agent therefore identifies potential confounders $Z$ among covariates strongly associated with $T^\star$ and recomputes the effect after adjustment. If the relationship disappears after adjustment, the initial evidence is not sufficient to support the proposed claim. Otherwise, the evidence collected during refinement is used to specify the variables, conditions, direction, and magnitude of the final hypothesis before it is added to the accepted set. Refer to Appendix~\ref{app:refinement} for implementation details.

\vspace{-0.3cm}
\section{Experiments}
\label{sec:experiments}
\vspace{-0.3cm}

We evaluate four aspects of \task: \textbf{RQ1:} How accurately and efficiently do methods discover hypotheses? \textbf{RQ2:} How does hypothesis structure affect recovery? \textbf{RQ3:} How do data size and task complexity affect discovery? \textbf{RQ4:} What does each component of \agent{} contribute?

\noindent\textbf{Empirical Setup.}
We compare \agent{} with eight baselines spanning statistical and causal methods (\textbf{Correlation Mining}, \textbf{Subgroup Analysis}, and \textbf{PC}~\citep{Spirtes1993CausationPA}), LLM-based methods (\textbf{XGBoost+LLM} and \textbf{LLM (no code)}), and agents (\textbf{ReAct}~\citep{react}, \textbf{Claude Code}, and \textbf{AutoDiscovery}~\citep{agarwal2025autodiscovery}). All LLM-based methods use Claude Sonnet 4.6, and all outputs are scored using the same \benchmark{} evaluator with 60{,}000 post-intervention samples and a $10\%$ effect-magnitude tolerance. We compute confidence intervals using 2{,}000 cluster-bootstrap resamples over tasks. Appendix~\ref{app:baseline_implementations} details method inputs, configurations, and output adapters.
\vspace{-0.3cm}
\subsection{RQ1: Overall Performance on \task}

\label{subsec:rq1}

\begin{figure}[H]
\centering
\begin{minipage}[t]{0.51\textwidth}
\vspace{0pt}
\centering
\textbf{(a) Overall performance and runtime}\par\medskip
\resizebox{\linewidth}{!}{%
\begin{tabular}{lcccc}
\toprule
Method & Precision & Recall & $F_1$ & Time (s/task) \\
\midrule
Correlation Mining & 0.00 & 0.01 & 0.00 & 53 \\
Subgroup Analysis  & 0.06 & 0.05 & 0.05 & 158 \\
PC                 & 0.00 & 0.00 & 0.00 & 504 \\
\midrule
XGBoost+LLM        & 0.20 & \underline{0.40} & 0.30 & 778 \\
LLM (no code)      & 0.18 & 0.33 & 0.24 & 711 \\
\midrule
AutoDiscovery      & 0.21 & 0.38 & 0.27 & 2{,}458 \\
ReAct              & 0.09 & 0.16 & 0.11 & 1{,}986 \\
Claude Code        & \underline{0.31} & 0.39 & \underline{0.35} & 3{,}308 \\
\midrule
\agent{}           & \textbf{0.47} & \textbf{0.51} & \textbf{0.49} & 2{,}681 \\
\bottomrule
\end{tabular}%
}
\end{minipage}\hfill
\begin{minipage}[t]{0.45\textwidth}
\vspace{-0.4    cm}
\centering
\textbf{(b) Recall across proposal budgets}\par\medskip
\includegraphics[width=\linewidth]{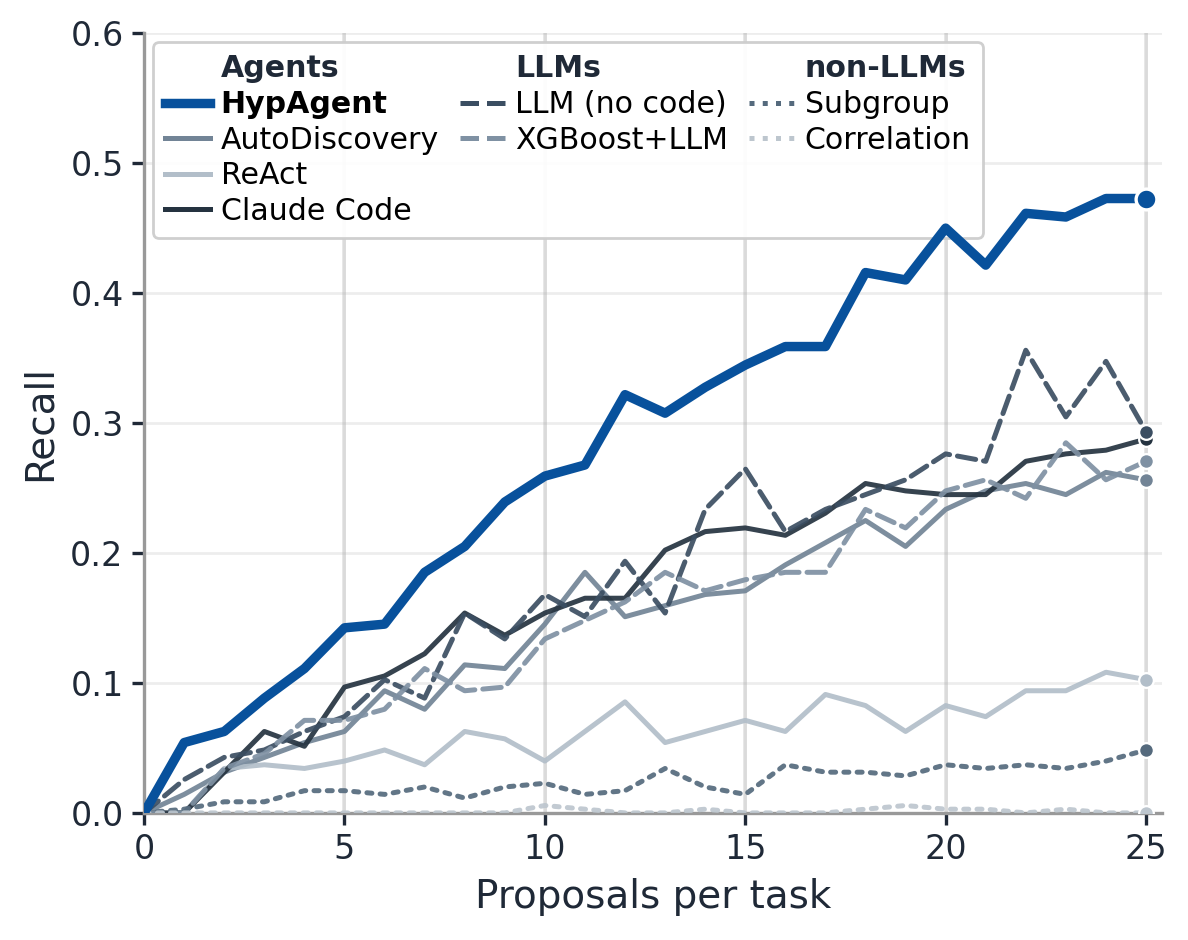}
\end{minipage}
\vspace{-0.3cm}
\caption{Overall discovery performance on \benchmark.
(a) Point estimates at each method's reported operating point; time is mean end-to-end discovery time per task and excludes the shared evaluator. Confidence intervals are reported in
Figure~\ref{app:main_confidence_intervals} in the appendix.
(b) Recall after uniformly subsampling each method to a common budget of up to 25 proposals per task.}
\label{fig:main_results}
\end{figure}
\vspace{-0.5cm}
\textbf{\agent{} achieves the strongest overall discovery performance.} Figure~\ref{fig:main_results} compares methods on precision, recall, and $F_1$. \agent{} achieves a precision of 0.47, recall of 0.51, and $F_1$ of 0.49. Claude Code has the strongest baseline precision at 0.31 and the strongest baseline $F_1$ at 0.35, while XGBoost+LLM has the strongest baseline recall at 0.40. \agent{} therefore improves both the correctness and coverage of the discovered hypotheses.

 \textbf{Failure modes vary across method families.} Statistical and causal methods achieve near-zero performance. These tools are effective when given a prespecified treatment or goal, but lack a mechanism to identify the conditions and effect magnitudes required by \benchmark. LLM-based methods generate a wider range of hypotheses, but their higher recall is accompanied by lower precision. In particular, we observe that the vanilla LLM baseline can propose plausible hypotheses but cannot test them against the data, while XGBoost+LLM grounds its proposals in feature importance without fully validating their causal structure. These results show that \task requires both expressive hypothesis generation and data-grounded validation. Agentic baselines show contrasting performance, with ReAct performing the worst while Claude Code is the strongest baseline agent by $F_1$. AutoDiscovery's surprisal objective may help explain its similar performance to generic LLM methods because an unexpected observational association can receive more search priority than a valid but unsurprising causal mechanism. This is consistent with its relatively high recall (0.38) but modest precision (0.21) on \benchmark.

\textbf{\agent{} remains strongest under matched proposal budgets.} Figure~\ref{fig:main_results}b further compares recall under matched proposal budgets. \agent achieves the highest recall throughout the range from one to 25 proposals per task, with its advantage generally increasing as the budget grows. This result indicates that \agent provides a better precision-recall tradeoff rather than obtaining greater coverage through a larger output set. Other LLM-based methods broadly follow similar trajectories. This suggests that the increase in recall of LLMs using XGBoost as a feature selector stems from their greater quantity of hypothesis proposals. Finally, \textbf{\task remains far from solved.} No baseline exceeds an $F_1$ of 0.35, and even \agent{} recovers only 51\% of the planted hypotheses.

\subsection{RQ2: Effect of Hypothesis Structure}

\label{subsec:rq2}

\begin{wrapfigure}[14]{r}{0.62\textwidth}
    \centering
    \includegraphics[width=0.60\textwidth]{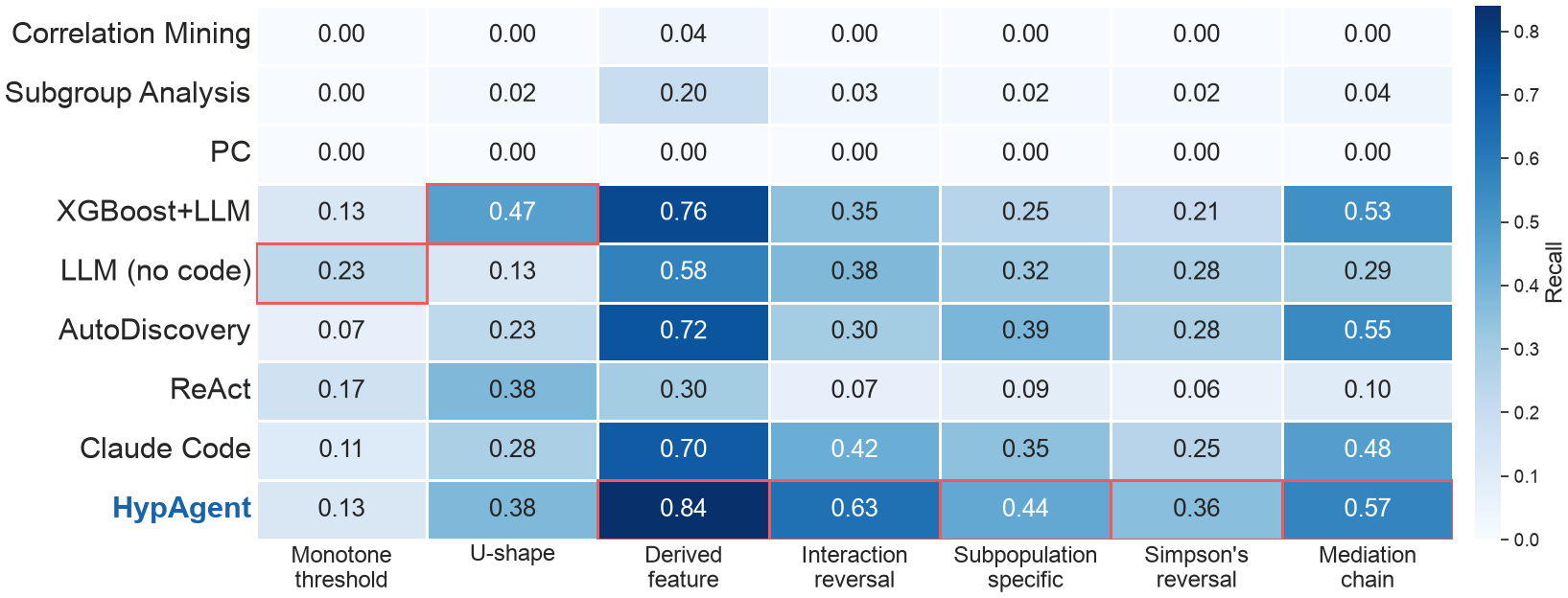}
    \caption{Recall across hypothesis archetypes with coral outlines marking the best-performing method for each archetype.}
    \label{fig:archetype_recall}
\end{wrapfigure}
\textbf{Hypothesis structure strongly determines discoverability.} 
Figure~\ref{fig:archetype_recall} shows that discovery performance varies substantially across hypothesis structures. \agent{} achieves the highest recall on five of the seven archetypes, whereas XGBoost+LLM performs best on U-shaped effects and LLM (no code) performs best on monotone effects. These structures require more than selecting the correct variable: a threshold claim must identify where the relationship changes, a subpopulation claim must specify where it applies, and a Simpson's-reversal claim must distinguish the marginal direction from the direction within strata.

\textbf{Compositional mechanisms are easier to recover than thresholds and aggregation reversals.}
We observe that hypotheses formed as a derivation of other features are consistently easier to discover, with performance between 0.58 and 0.84 among the stronger evaluated methods. Surprisingly, monotone threshold hypotheses are difficult to recover across all methods, as are Simpson's reversals and hypotheses specific to a subpopulation. U-shaped and interaction-reversal hypotheses exhibit higher variance in performance, while hypotheses mediated or confounded by related features are modestly recovered by agentic baselines. Overall, methods perform better on hypotheses involving semantically meaningful feature compositions or causal chains than on hypotheses whose structure depends on subtle thresholds, subgroup boundaries, or reversals under aggregation.

\textbf{The main source of failure is characterizing the effects of relevant variables.} We systematically analyze \agent's failed recoveries by categorizing the 172 missed planted hypotheses. Wrong-form errors account for 87 misses (50.6\%), missed variables account for 53 (30.8\%), and wrong effect magnitudes account for 32 (18.6\%). Under these criteria, characterization errors account for 119 of the 172 missed mechanisms (69.2\%), after a relevant variable has already been identified. Refer to Figure~\ref{fig:failure_modes} and Appendix~\ref{app:failure_analysis} for the complete breakdown.

\subsection{RQ3: Factors Affecting \task}
\label{subsec:rq3}

\begin{figure}[H]
    \centering
    \includegraphics[width=\textwidth]{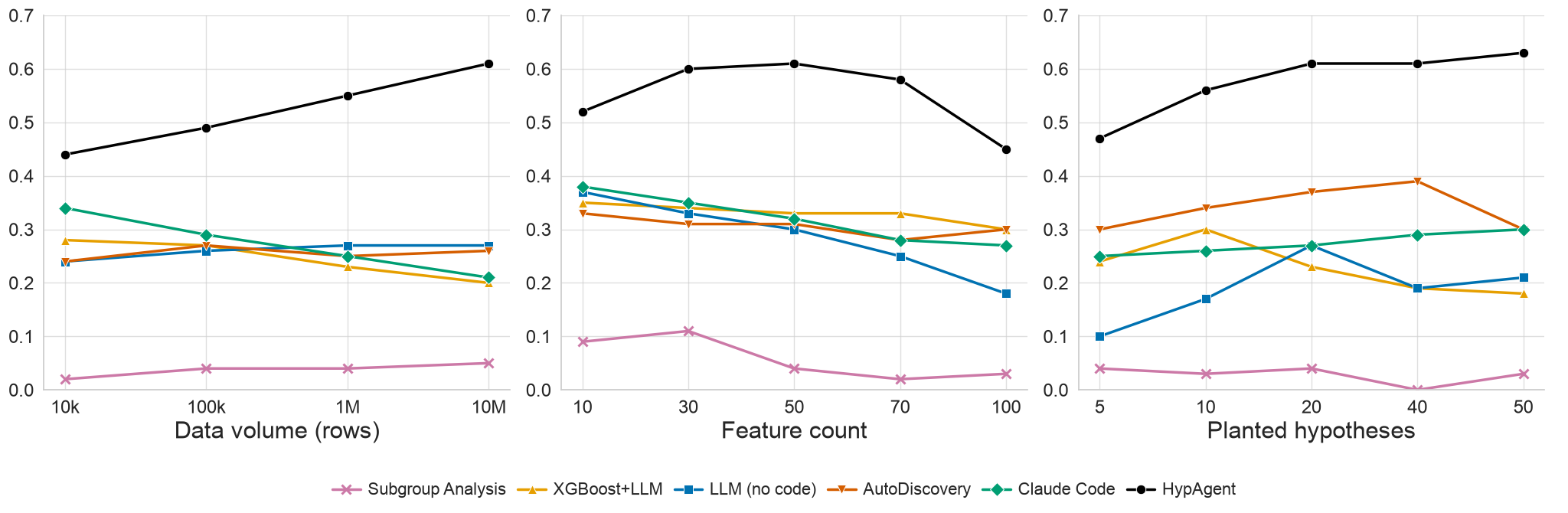}
    \vspace{-0.6cm}
    \caption{$F_1$ as the number of rows, features, and planted hypotheses varies.}
    \label{fig:scaling}
\end{figure}
\vspace{-0.5cm}
To study the factors affecting discovery, we vary one benchmark dimension at a time across all domains while holding the others at a common default of $10^6$ rows, 50 features, and 20 planted hypotheses, and average the results across tasks. We observe that additional data benefit \agent{}, whose $F_1$ increases from 0.44 at $10^4$ rows to 0.61 at $10^7$, while most baselines remain flat or decline. Increasing the number of features generally hurts performance as \agent{} reaches $F_1=0.61$ at 50 features but falls to 0.45 at 100. By contrast, increasing the number of planted hypotheses does not consistently increase difficulty as \agent{} improves from $F_1=0.47$ with five hypotheses to approximately 0.63 with 50, while the baselines exhibit weaker or inconsistent trends. These findings indicate that more data strengthens the evidence available for testing a candidate, and more planted mechanisms can provide additional discoverable signal. Hence, \textbf{the search space matters more than the amount of available evidence}, since \task provides no prespecified target and its main difficulty lies in deciding which explanations to test rather than gathering enough observations to test them.
\vspace{-0.2cm}
\subsection{RQ4: Contribution of \agent Components}
\label{subsec:rq4}
\vspace{-0.2cm}
\begin{wraptable}{r}{0.7\textwidth}
    \centering
    \vspace{-0.6cm}
    \caption{Ablation of \agent components across all 15 tasks. Ablated rows
    report changes relative to the full method.}
    \label{tab:ablations}
    \scriptsize
    \setlength{\tabcolsep}{3pt}
    \resizebox{\linewidth}{!}{%
    \begin{tabular}{@{}lcccl@{}}
        \toprule
        \textbf{Configuration} & \textbf{P} & \textbf{R} &
        $\mathbf{F_1}$ & \textbf{Impact} \\
        \midrule
        \agent{}                  & 0.47  & 0.51  & 0.49  & -- \\
        \quad $-$ Residualization & $-0.15$ & $-0.12$ & $-0.13$ &
        68.8\% increase in duplicates \\
        \quad $-$ Triage          & $-0.08$ & $-0.13$ & $-0.11$ &
        Suboptimal direction in 25.5\% of cases \\
        \quad $-$ Refinement      & $-0.31$ & $-0.36$ & $-0.34$ &
        58.5\% increase in unsupported hypotheses \\
        \bottomrule
    \end{tabular}%
    }
\end{wraptable}

We ablate each component of \agent{} across all 15 tasks of \benchmark. Our findings reveal that residualization reduces repeated attention to dominant explanations, allowing later iterations to focus on unexplained outcome variation and reducing duplicate hypotheses. Triage directs the limited refinement budget toward candidate directions with stronger residual evidence. Refinement has the largest contribution because it transforms a broad direction into a specific hypothesis by determining its conditions, effect, and relevant variables, while filtering observational patterns that are not robust. Together, the components support complementary stages of discovery: residualization promotes diversity, triage improves search allocation, and refinement improves the specificity and reliability of the resulting hypotheses.

\vspace{-0.3cm}
\section{Conclusion}
\label{sec:conclusion}
\vspace{-0.3cm}
\task{} shifts the focus from explaining a prespecified relationship to discovering multiple data-supported hypotheses from a dataset and an outcome. We introduced \benchmark{} to evaluate this capability in controlled causal worlds, separating observational discovery from interventional validation, and \agent{} to address the problem. \agent{} outperforms all baselines, yet most of its missed recoveries occur after a relevant variable has already been surfaced. Progress in this setting therefore depends less on generating more plausible proposals and more on deciding where to search and precisely characterizing how a discovered relationship behaves. We hope \benchmark{} and \agent{} serve as a foundation for studying \task and enable broader discovery systems. Appendix~\ref{sec:limitations} discusses the scope and limitations of this study.

\newpage

\subsubsection*{Ethics statement}
\benchmark{} uses synthetic observations generated from structural causal models grounded in the schemas and empirical marginal distributions of publicly available datasets. This study does not recruit human participants, use private datasets, or perform interventions on real-world populations. The synthetic mechanisms should not be interpreted as established causal relationships in the source populations, and we do not claim a formal privacy guarantee from synthetic generation alone. Source datasets and their listed licenses are documented in Appendix~\ref{app:benchmark_details}, Table~\ref{tab:all_tasks}. The principal risks are overstating autonomous discovery capabilities and treating simulated causal support as sufficient for real-world decisions. Accordingly, \agent{} produces candidate hypotheses for further investigation, not actionable scientific or policy recommendations; real-world use requires independent, domain-appropriate validation.


\subsubsection*{Reproducibility statement}
Benchmark construction is described in Section~\ref{sec:benchmark} and Appendix~\ref{app:benchmark_details}; interventional evaluation and recall matching are described in Section~\ref{sec:interventional_evaluation} and Appendix~\ref{app:evaluation_protocol}. Agent configurations and baseline implementations are provided in Appendices~\ref{app:hypagent_configuration} and~\ref{app:baseline_implementations}, respectively. For future evaluations, fresh discovery targets require re-instantiating mechanisms and their parameters rather than merely resampling observations from a published SCM.



\bibliography{iclr2027_conference}
\bibliographystyle{iclr2027_conference}

\appendix


\section{Limitations}
\label{sec:limitations}

This work focuses on controlled evaluation of open-ended discovery in large tabular datasets. \benchmark{} uses synthetic causal mechanisms grounded in real-world schemas, enabling precise interventional evaluation without treating the resulting mechanisms as verified relationships in the source domains. We study binary outcomes and seven mechanism families, while extensions to other outcomes, modalities, and forms of scientific explanation remain outside the present scope. To isolate differences between discovery procedures, all LLM-based methods use a common backbone. Cross-family evaluation is left to future work. Similarly, \agent{} performs observational discovery and is evaluated through interventions within the benchmark SCM, rather than claiming causal identification in real-world applications. Our evaluation measures interventional support and mechanism recovery under a finite search budget. Assessing downstream scientific utility with domain experts and constructing continually refreshed hidden mechanisms are important directions for future evaluation.

\section{Functional Form Definitions}
\label{app:functional_forms}

Let $T$ denote a treatment variable, $M$ a moderator or subgroup variable,
and $Y$ the binary outcome. Thresholds and subgroup boundaries are fixed from
the empirical quantiles of the corresponding task variables at generation
time. Table~\ref{tab:functional_forms} summarizes the seven structural forms
represented in \benchmark.

\begin{table}[ht]
\centering
\caption{Functional forms in \benchmark. Counts are aggregated across all
15 tasks.}
\label{tab:functional_forms}
\small
\setlength{\tabcolsep}{4pt}
\begin{tabular}{p{2.5cm}p{8.4cm}r}
\toprule
Form & Definition and interventional signature & Count \\
\midrule
Monotone threshold
& $T$ changes $Y$ after crossing a threshold, e.g.,
  $\phi(T)=w\mathbb{I}[T>q]$. & 30 \\
Non-monotone
& The effect is strongest at both tails or within an interior band; the
  resulting curve contains at least one direction change. & 47 \\
Derived feature
& The treatment is a function of multiple variables,
  $U=g(T_1,T_2)$, such as a ratio, product, or difference. & 50 \\
Interaction reversal
& The direction of the effect of $T$ changes across values of a moderator variable
  $M$. & 71 \\
Subpopulation-specific
& The effect is active only under a condition $c(M)$. & 57 \\
Simpson's reversal
& The marginal association between $T$ and $Y$ has the opposite direction
  from the association within strata of $M$. & 47 \\
Mediation
& The treatment affects the outcome through an intermediate variable,
  $T\rightarrow M\rightarrow Y$. & 49 \\
\midrule
\textbf{Total} & & \textbf{351} \\
\bottomrule
\end{tabular}
\end{table}

\section{Benchmark Details}
\label{app:benchmark_details}

\benchmark contains 15 tasks spanning census, e-commerce, education,
finance, and health. Each task borrows its variable vocabulary, types,
semantics, and empirical marginal distributions from a public dataset. The
causal graph, structural equations, planted mechanisms, and generated rows
are constructed specifically for \benchmark. Table~\ref{tab:all_tasks}
reports the complete task inventory. Every outcome is binary; the outcome
column describes the event represented by $Y=1$.

\begin{table*}[t]
\centering
\caption{Task list of \benchmark.}
\label{tab:all_tasks}
\scriptsize
\setlength{\tabcolsep}{3pt}
\resizebox{\textwidth}{!}{
\begin{tabular}{lllcllp{3.1cm}rl}
\toprule
Task ID & Domain & Source dataset & URL & License & Rows & Outcome ($Y=1$)
& Features & Hyp. \\
\midrule
\texttt{census\_acs} & Census
& ACS PUMS~\citep{census_acs_pums}
& \href{https://www.census.gov/programs-surveys/acs/microdata.html}{link}
& U.S. public domain & 1M & Housing cost burden & 63 & 23 \\
\texttt{census\_adult} & Census
& UCI Adult~\citep{uci_adult}
& \href{https://archive.ics.uci.edu/dataset/2/adult}{link}
& CC BY 4.0 & 5M & High income & 63 & 25 \\
\texttt{census\_cps} & Census
& CPS ASEC~\citep{census_cps_asec}
& \href{https://www.census.gov/data/datasets/time-series/demo/cps/cps-asec.html}{link}
& U.S. public domain & 10M & Long commute & 64 & 25 \\
\midrule
\texttt{ecom\_hm} & E-commerce
& H\&M~\citep{kaggle_hm}
& \href{https://www.kaggle.com/competitions/h-and-m-personalized-fashion-recommendations/data}{link}
& Competition rules & 5M & Customer churn & 61 & 23 \\
\texttt{ecom\_instacart} & E-commerce
& Instacart~\citep{kaggle_instacart}
& \href{https://www.kaggle.com/competitions/instacart-market-basket-analysis/data}{link}
& Competition rules & 1M & Product reordered & 63 & 21 \\
\texttt{ecom\_onlineretail} & E-commerce
& Online Retail II~\citep{uci_online_retail_ii}
& \href{https://archive.ics.uci.edu/dataset/502/online+retail+ii}{link}
& CC BY 4.0 & 10M & High lifetime value & 62 & 24 \\
\midrule
\texttt{edu\_oulad} & Education
& OULAD~\citep{uci_oulad}
& \href{https://archive.ics.uci.edu/dataset/349/open+university+learning+analytics+dataset}{link}
& CC BY 4.0 & 1M & Course completion & 58 & 22 \\
\texttt{edu\_studentperf} & Education
& Student Performance~\citep{uci_student_performance}
& \href{https://archive.ics.uci.edu/dataset/320/student+performance}{link}
& CC BY 4.0 & 5M & High exam performance & 64 & 24 \\
\texttt{edu\_xapi} & Education
& xAPI-Edu-Data~\citep{kaggle_xapi_edu}
& \href{https://www.kaggle.com/datasets/aljarah/xAPI-Edu-Data}{link}
& CC BY-SA 4.0 & 10M & High engagement class & 65 & 24 \\
\midrule
\texttt{fin\_givemecredit} & Finance
& Give Me Some Credit~\citep{kaggle_give_me_some_credit}
& \href{https://www.kaggle.com/competitions/give-me-some-credit/data}{link}
& Competition rules & 5M & Financial distress & 58 & 25 \\
\texttt{fin\_lendingclub} & Finance
& Lending Club~\citep{kaggle_lending_club}
& \href{https://www.kaggle.com/datasets/wordsforthewise/lending-club}{link}
& CC0 listing & 1M & High interest-rate tier & 65 & 24 \\
\texttt{fin\_taiwan} & Finance
& Taiwan Default~\citep{uci_taiwan_default}
& \href{https://archive.ics.uci.edu/dataset/350/default+of+credit+card+clients}{link}
& CC BY 4.0 & 10M & Payment default & 59 & 21 \\
\midrule
\texttt{health\_brfss} & Health
& BRFSS~\citep{cdc_brfss}
& \href{https://www.cdc.gov/brfss/annual_data/annual_data.htm}{link}
& U.S. public domain & 5M & Frequent poor mental health & 65 & 21 \\
\texttt{health\_diabetes130} & Health
& Diabetes 130-Hospitals~\citep{uci_diabetes_130}
& \href{https://archive.ics.uci.edu/dataset/296/diabetes+130-us-hospitals-for-years-1999-2008}{link}
& CC BY 4.0 & 10M & Readmission within 30 days & 53 & 24 \\
\texttt{health\_nhanes} & Health
& NHANES~\citep{cdc_nhanes}
& \href{https://wwwn.cdc.gov/nchs/nhanes/}{link}
& NCHS public-use data agreement & 1M & Metabolic syndrome & 63 & 25 \\
\midrule
\multicolumn{8}{r}{\textbf{Total planted hypotheses}} & \textbf{351} \\
\bottomrule
\end{tabular}}
\end{table*}

Overall, the benchmark contains 53--65 visible features and 1M--10M rows per
task. The three tasks in each domain use distinct source schemas rather than
different samples from the same schema.

\subsection{Hypothesis and SCM Construction}
\label{app:scm_construction}

Candidate mechanisms were proposed using Claude Opus 4.8 through the
Anthropic API. For each task, the model received the domain, the current
schema, the designated outcome, and four few-shot examples. It did not receive observations from the generated task or the final SCM. Each curated mechanism was represented as $h=(X_h,\psi,d,c,m_h)$ and translated into an individual causal DAG. Here, $X_h$ identifies the variables involved, $\psi$ defines the structural form, $d$ gives the direction, $c$ specifies an optional population condition, and $m_h$ is the absolute interventional change in outcome probability. If a given hypothesis could be cleanly represented as a tuple, we retained it. We treat two hypotheses as distinct when they involve different participating-variable sets $X_h$, different structural forms, or non-equivalent conditions. Differences in wording or effect-magnitude estimates alone do not constitute distinct mechanisms. These hypotheses are then converted to DAGs and composed into one task-level graph, which defines the SCM.

Variables may serve as background causes, confounders, moderators, mediators,
causes, decoys, or null variables. Background variables are upstream causes
without outcome terms. Confounders cause both a treatment and the outcome,
moderators alter another mechanism through an interaction or gate, and
mediators lie on a directed treatment--outcome path. Causes correspond to
mechanisms in the answer key. Decoys correlate with causes through shared
parents but have no outcome term, while null variables are independent.

Before materialization, we reject specifications containing undefined
parents, directed cycles, missing moderators, or numerical primitives applied
to categorical variables. This guarantees a valid topological generation
order for every task.

\section{Evaluation Protocol}
\label{app:evaluation_protocol}

Each proposal contains its natural-language claim, involved variables, claimed direction and magnitude, and an optional condition. The seven archetypes describe the mechanisms planted in \benchmark and hypothesis generators are not required to name these categories. LLM-based methods return the required fields as structured outputs, while fixed adapters convert statistical and causal baseline outputs into the same representation. We first parse and validate these fields deterministically. If parsing fails, Claude Opus 4.8 performs a fallback parse using fixed few-shot examples, temperature zero, and JSON output. Proposals that remain invalid are counted as unsupported. In our evaluations, none of the proposed hypotheses had any parsing issues.

Before evaluation, proposals are deduplicated using the mechanism-identity rule
in Section~\ref{sec:problem}, retaining the first proposal encountered for each
identity. Precision is the fraction of these distinct proposals supported by
intervention. For
recall, supported proposals are first shortlisted against planted mechanisms
using variable overlap, interventional response shape, condition, and
magnitude. Claude Opus 4.8 then makes the final equivalence decision at
temperature zero with JSON output. Multiple proposals matched to the same
mechanism count once. A supported proposal that matches no planted mechanism
contributes to precision but not recall, allowing valid hypotheses outside the planted set to receive credit.

\subsection{\agent{} Implementation}
\label{app:hypagent_implementation}

\subsubsection{Residualization}
\label{app:residualization}

After every accepted hypothesis, residualization is updated using all variables named by the accepted hypotheses. Each continuous variable is expanded into indicators for its five empirical-quantile bins. Each categorical variable is one-hot encoded. 
Let $\Phi(S_k)$ denote the combined representation of the hypotheses accepted
through round $k$. We fit an ordinary least-squares model to the original
binary outcome:
\begin{equation}
    \hat f_k
    =
    \arg\min_f
    \left\|Y-f\!\left(\Phi(S_k)\right)\right\|_2^2,
    \qquad
    \widetilde Y_k
    =
    Y-\hat f_k\!\left(\Phi(S_k)\right).
\end{equation}
The fit uses the full dataset and is jointly recomputed after every acceptance. Overlapping hypotheses are therefore not subtracted independently. Their shared explanatory contribution is resolved by the joint fit. Rejected hypotheses remain in memory but do not enter $\Phi(S_k)$.

\subsubsection{Proposal and Triage}
\label{app:proposal_triage}

At each round, the proposer generates six broad candidate directions for explaining the current residual based on a randomly queried slice of data. Its context contains the dataset schema, outcome name, current residual, accepted and rejected hypothesis memories, and results returned by previous analysis calls. The context is assembled
from these elements at each round rather than retaining an unrestricted conversation history. The LLM accesses data by writing analysis code and receives the resulting statistics or summaries rather than the complete table in its context.
Triage evaluates each candidate through a candidate-specific query of 1,000 observations. 
The triage score is
\begin{equation}
s(T)=
\frac{
    \operatorname{Var}\!\left(
        \mathbb{E}[\widetilde Y_k\mid\operatorname{bin}(T)]
    \right)
}{
    \operatorname{Var}(\widetilde Y_k)
}.
\end{equation}
This score measures the fraction of residual variation associated with the candidate. Candidates with $s(T)<0.02$ are discarded. The highest-scoring remaining direction is passed to refinement.

If none of the six candidates clears the threshold, the search terminates.
Residual fitting and refinement use the full dataset.

\subsubsection{Refinement}
\label{app:refinement}

Refinement receives the selected direction, dataset schema, outcome, full observational dataset, and memories of previously accepted and rejected hypotheses. It may execute arbitrary analyses for up to 10 code turns. The LLM is not given a fixed menu of functional forms. It determines how to
formulate the relationship from its analysis and may select thresholds, ranges, moderators, treatment values, conditions, and derived features
$U=g(T_1,T_2)$.

Derived features are introduced during refinement after triage has selected an initial direction. The LLM chooses both the constituent variables and the function $g$ from its data analysis. Finally, \agent examines the 15 variables most
correlated with the selected candidate. Continuous moderators retain their numerical values and categorical moderators are one-hot encoded. A moderator is retained when its interaction coefficient satisfies $p<0.01$ and its
standardized magnitude exceeds the fixed hyperparameter $0.005$. 

The final natural-language hypothesis explicitly states its involved variables, direction $d$, magnitude $m_h$, and any condition $c$. It does not return a separate benchmark functional-form label. The relationship described
in the hypothesis text is mapped to the benchmark representation only during evaluation.

Direction and magnitude are estimated from the original binary outcome rather than from the residual used during triage. The LLM selects the treatment values according to its analysis. The claimed magnitude is
\begin{equation}
    m_h
    =
    \left|
        \widehat{\Pr}(Y=1\mid T=t_{\mathrm{hi}},c)
        -
        \widehat{\Pr}(Y=1\mid T=t_{\mathrm{lo}},c)
    \right|.
\end{equation}
When $c$ is absent, the contrast is calculated over the full population.

Before acceptance, the claimed direction is recomputed from the data and must agree with the direction stated in the hypothesis:
\begin{equation}
    \operatorname{sign}(\hat\tau_{\mathrm{data}})
    =
    \operatorname{sign}(\hat\tau_{\mathrm{claim}}).
\end{equation}
For $d=\mathrm{mixed}$, this agreement is checked separately within each region or condition stated by the hypothesis rather than through one aggregate sign.
The effect must also survive covariate adjustment. We select the 12 covariates most correlated with the proposed treatment and include them jointly in an ordinary least-squares regression of the current residual
$\widetilde Y_k$ on the treatment and selected covariates. Let $\hat\beta_T^{\mathrm{adj}}$ denote the adjusted treatment coefficient. We convert it to an interdecile effect using
\begin{equation}
    \Delta_T
    =
    q_{0.9}(T)-q_{0.1}(T),
    \qquad
    \hat\tau_{\mathrm{adj}}
    =
    \hat\beta_T^{\mathrm{adj}}\Delta_T.
\end{equation}
For a conditional hypothesis, this calculation is performed on the subset satisfying its stated condition. A hypothesis is accepted when its direction agrees with the recomputed direction and \begin{equation}
    \left|\hat\tau_{\mathrm{adj}}\right|>0.02.
\end{equation}
Because $Y$ is binary, this threshold corresponds to an adjusted two-percentage-point outcome difference over the treatment's interdecile range. 

Every refinement attempt consumes one unit of the budget $B=50$, regardless of whether the resulting hypothesis is accepted. A rejected complete claim is stored in memory and is not retried. Exclusion applies to the complete
claim rather than its variables. This allows subsequent hypotheses to reuse a variable under a different relationship or condition.

Although the budget permits up to 50 refinement attempts, it did not bind in
the reported runs. Runs terminated after 20--36 refinement attempts when no
candidate in the current proposal batch cleared the triage threshold.

\subsection{\agent Configuration}
\label{app:hypagent_configuration}

All LLM-based discovery methods use Claude Sonnet 4.6. \agent proposes six
candidate directions per round at temperature 0.7, scans moderators at
temperature 0.6, and validates at temperature 0.5. It uses five quantile bins
for continuous variables and observed categories for categorical variables.
The triage threshold is $\epsilon=0.02$. For confounding checks in refinement, \agent
examines the 12 observed covariates most strongly associated with the
candidate treatment. A proposal is rejected when its absolute interdecile-scaled
adjusted effect is at most 0.02. Runs terminate when no candidate clears
the triage threshold. In the reported runs, \agent used 20--36 iterations, with a mean of 26.3.

\begin{algorithm}[t]
\caption{\agent: \task through iterative residual search.}
\label{alg:hypagent}
\begin{algorithmic}[1]

\REQUIRE Dataset $D$, outcome $Y$, refinement budget $B$,
triage threshold $\epsilon$
\ENSURE Accepted hypothesis set $S$

\STATE $S\gets\emptyset$
\COMMENT{accepted hypotheses}
\STATE $S_{\mathrm{rej}}\gets\emptyset$
\COMMENT{rejected complete claims}

\WHILE{budget $B$ remains}

    \STATE $\widetilde Y\gets\textsc{Residualize}(D,Y,S)$
    \COMMENT{remove variation represented by accepted hypotheses}

    \STATE $\mathcal C\gets
    \textsc{Propose}(\mathrm{schema}(D),\widetilde Y,S,S_{\mathrm{rej}})$
    \COMMENT{propose six candidate directions}

    \STATE $T^\star\gets
    \textsc{Triage}(D,\widetilde Y,\mathcal C,\epsilon)$
    \COMMENT{select using residual signal}

    \IF{$T^\star=\varnothing$}
        \STATE \textbf{break}
        \COMMENT{no candidate clears triage}
    \ENDIF

    \STATE $\hat h\gets
    \textsc{Refine}(D,\widetilde Y,T^\star)$
    \COMMENT{formulate and validate the claim}

    \STATE \textsc{ConsumeBudget}$(B)$
    \COMMENT{accepted and rejected attempts count}

    \IF{$\hat h$ passes the acceptance checks}
        \STATE $S\gets S\cup\{\hat h\}$
        \COMMENT{update the residual in the next round}
    \ELSE
        \STATE $S_{\mathrm{rej}}\gets S_{\mathrm{rej}}\cup\{\hat h\}$
        \COMMENT{do not retry the same complete claim}
    \ENDIF

\ENDWHILE

\RETURN $S$

\end{algorithmic}
\end{algorithm}
\subsection{Baseline Implementations}
\label{app:baseline_implementations}

\textbf{Correlation Mining} ranks features by point-biserial correlation with the outcome and
returns the leading features as monotone hypotheses in the observed
direction. \textbf{Subgroup Analysis} searches single-feature subgroups and
ranks them by outcome lift. \textbf{PC} applies conditional-independence tests
and returns edges adjacent to the outcome as candidate causal relationships.
\textbf{XGBoost+LLM} fits a gradient-boosted classifier and supplies its
ranked feature importances, variable-distribution summaries, and 1{,}000
sampled rows to the LLM, which formulates complete hypotheses. Following the
example-driven setup of \citet{zhou2024hypothesis}, \textbf{LLM (no code)}
receives the domain, schema, outcome, variable-distribution summaries, and
1{,}000 sampled rows, but cannot execute code or request additional data.
\textbf{ReAct} interleaves reasoning with code
execution. \textbf{Claude Code} may write and execute analysis code over the
complete table. \textbf{AutoDiscovery} proposes and tests discoveries using
its original iterative procedure. All returned hypotheses are processed by
the shared parser and scored by the same interventional evaluator. All LLM-based methods use Claude Sonnet 4.6 as the backbone.

\begin{figure}[t]
\centering
\includegraphics[width=\textwidth]{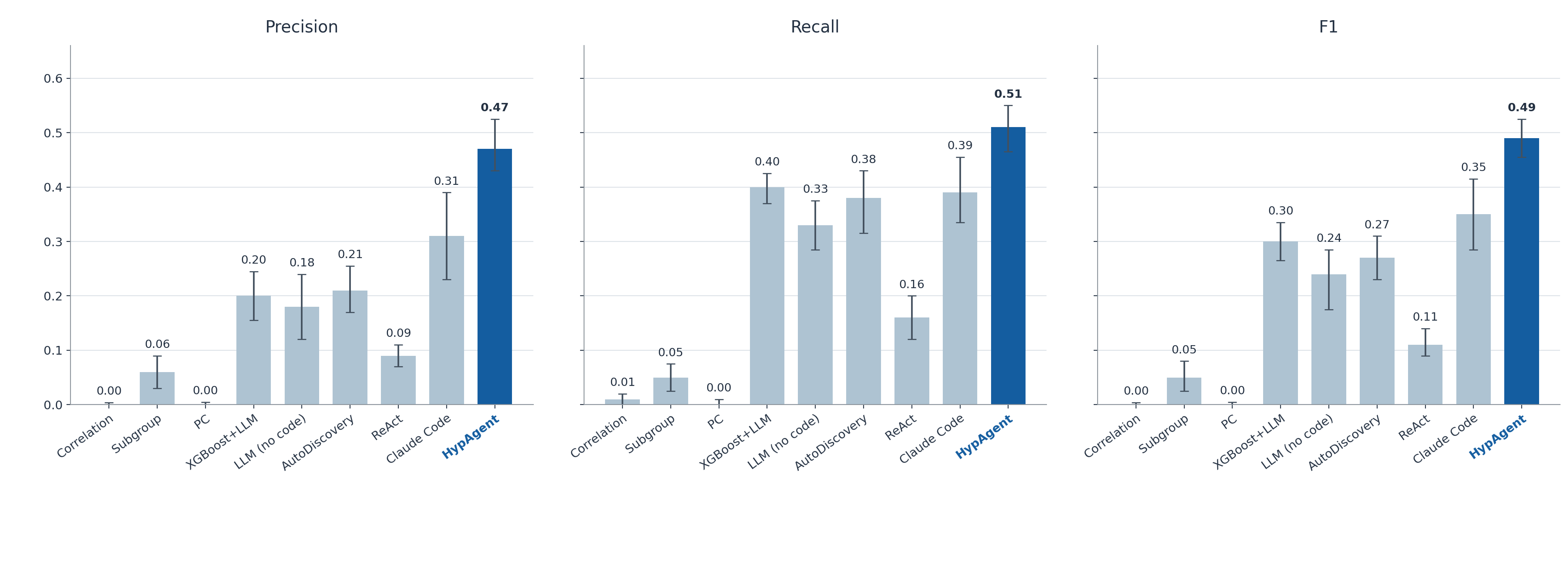}
\caption{Precision, recall, and $F_1$ at each method's reported operating
point. Whiskers show 95\% cluster-bootstrap confidence intervals over tasks.}
\label{app:main_confidence_intervals}
\end{figure}

\subsection{Failure Analysis}
\label{app:failure_analysis}

\begin{figure}[H]
    \centering
    \includegraphics[width=0.58\textwidth]{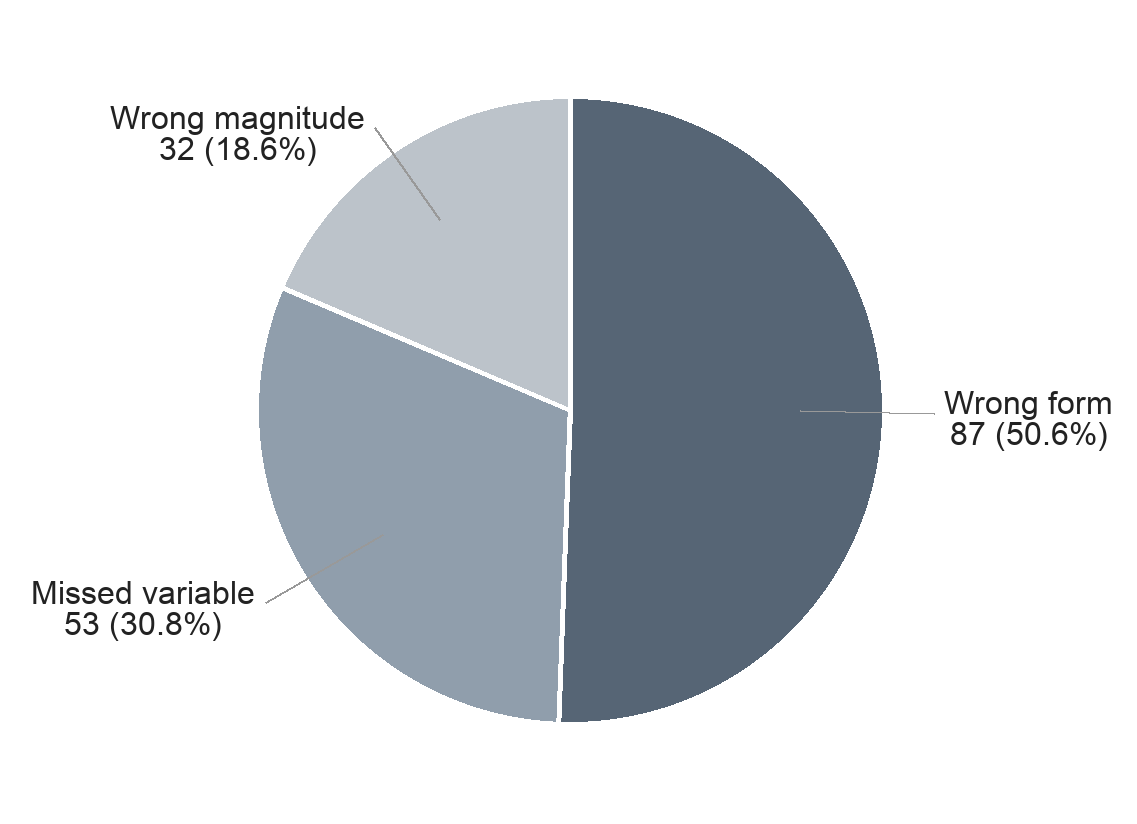}
    \caption{Failure modes among the 172 planted hypotheses missed by
    \agent{}.}
    \label{fig:failure_modes}
\end{figure}

\section{Prompt Templates}
\label{app:prompt_templates}

We report the prompts used to construct candidate benchmark mechanisms and for
the two LLM-driven stages of \agent. Triage, confounder adjustment, and final
validity checks are implemented programmatically and are not delegated to the
LLM.

\subsection{Benchmark Candidate Construction}
\label{app:benchmark_construction_prompt}

The following template is instantiated separately for each target functional
form. We request at least three candidates per form and retain only mechanisms
that pass the structural checks described in Appendix~\ref{app:scm_construction}.

\begin{promptbox}{Benchmark Candidate Construction}
\begin{verbatim}
SYSTEM:
You are an expert in causal modeling and synthetic benchmark
construction. Given a domain, dataset schema, outcome, and target
functional form, propose plausible causal hypotheses that can be
instantiated in a structural causal model. Return JSON only.

USER:
Domain:
{domain}

Dataset schema:
{schema}

Designated outcome:
{outcome}

Target functional form:
{functional_form}

Definition of the target functional form:
{functional_form_definition}

Examples:
{four_few_shot_examples}

Generate {k} distinct causal hypotheses that instantiate the specified
functional form. Every generated hypothesis must use that form.

Represent each hypothesis as h = (X_h, psi, d, c, m_h), where X_h
contains the variables involved, psi is the specified functional form,
d is the effect direction, c is an optional condition or
subpopulation,
and m_h is the expected effect magnitude under the relevant
intervention contrast.

Each hypothesis must:
- use only variables in the supplied schema;
- state an interventional causal claim rather than an association;
- identify the treatment and any moderator, mediator, stratum, or
  component variables required by the functional form;
- be precise enough to compile into a causal DAG and structural
  equation;
- be plausible in the supplied domain and substantively distinct from
  the other candidates;
- not use the outcome as an input variable; and
- not claim to have observed data patterns or statistics, because no
  dataset rows are provided.

Return a JSON list:
[
  {
    "X_h": ["<treatment>", "<optional additional variables>"],
    "psi": "{functional_form}",
    "relationship": "<precise causal relationship>",
    "d": "positive | negative | mixed",
    "c": "<condition or null>",
    "m_h": "<expected effect magnitude>",
    "causal_roles": {
      "<variable>": "treatment | moderator | mediator | stratum |
                      component"
    },
    "hypothesis": "<concise natural-language causal claim>"
  }
]
\end{verbatim}
\end{promptbox}

\subsection{\agent{} Prompts}

\begin{promptbox}{Candidate Proposal}
\begin{verbatim}
SYSTEM:
You are a data scientist proposing directions to investigate.
Return JSON only.

USER:
Dataset schema:
{schema}

Outcome:
{outcome}

Random sample of 1,000 rows:
{sample_rows}

Summary of the current residual:
{residual_summary}

Hypotheses already confirmed:
{confirmed_hypotheses}

Hypotheses previously rejected:
{rejected_hypotheses}

Results from previous analysis calls:
{previous_analysis_results}

Propose {k} distinct and promising new directions that could explain
variation in the outcome.

Each proposal should identify a primary variable to investigate and,
when relevant, other variables that may participate in the
relationship. Prioritize directions that are not already explained by
the confirmed hypotheses. Do not restate or superficially reformulate
a confirmed hypothesis.

At this stage, propose broad directions rather than complete
hypotheses. Do not assume a predefined functional form. A later
refinement step will determine the specific relationship, conditions,
direction, and effect magnitude.

Return a JSON list:
[
  {
    "variable": "<primary column>",
    "partners": ["<optional related column>", "..."],
    "rationale":
      "<brief reason this direction may explain the outcome>"
  }
]
\end{verbatim}
\end{promptbox}

\begin{promptbox}{Candidate Refinement}
\begin{verbatim}
SYSTEM:
You are a data scientist refining a promising direction into a
precise, testable hypothesis. Explore the data by writing Python.
Use <code>...</code> to run code and return the final result inside
<hypothesis>...</hypothesis> as JSON. Do not propose unrelated
directions.

USER:
Dataset schema:
{schema}

Outcome:
{outcome}

HYPOTHESES ALREADY CONFIRMED:
{confirmed_hypotheses}

HYPOTHESES PREVIOUSLY REJECTED:
{rejected_hypotheses}

DIRECTION SELECTED BY TRIAGE:
{candidate}

In your Python namespace you have:
- `df`: the full {n_rows:,} rows, containing all columns and the raw
  outcome.
- `resid`: a NumPy array aligned with `df`. It contains the outcome
  after the effects of all confirmed hypotheses have been removed
  and therefore represents the remaining unexplained signal.

Your job is to refine the selected direction into ONE precise
hypothesis. Focus on the selected primary variable and determine:
- which variables are involved;
- how they are related to the remaining outcome;
- whether the relationship depends on a range, condition, partner
  variable, or subpopulation;
- the direction of the relationship under the identified
  conditions; and
- its estimated magnitude.

Use code to inspect the selected variable across its distribution or
categories. Investigate partner variables or conditions only when
supported by the data. Compare the relevant groups or regions using
the full dataset before producing the final estimate.

Do not assume a predefined functional form. Let the observed
relationship determine the hypothesis. Do not switch to an unrelated
primary variable. Confounder adjustment and final validity checks are
performed separately after this step.

If the selected direction does not support a sufficiently specific
hypothesis, return:
<hypothesis>{"supported": false}</hypothesis>

Otherwise, return:
<hypothesis>
{
  "supported": true,
  "X_h": ["<primary variable>", "<other involved variables>"],
  "psi": "<concise description of the observed relationship>",
  "d": "<positive, negative, or mixed>",
  "c": "<condition, range, or subpopulation; null if unconditional>",
  "m_h": <estimated magnitude>,
  "text": "<one-sentence testable hypothesis>"
}
</hypothesis>

You have {max_turns} code turns.
\end{verbatim}
\end{promptbox}

\begin{promptbox}{LLM (no code)}
\begin{verbatim}
SYSTEM:
You are a data scientist generating testable hypotheses from
observed data. Return JSON only.

USER:
Domain: {domain}
Dataset description: {scenario}
Outcome: {outcome}

Schema:
{schema}

Variable distributions:
{distribution_summaries}

Random sample of 1,000 rows:
{sample_rows}

Using the observed examples and distribution summaries, generate
{k} distinct, testable hypotheses about variables that may affect
the outcome. You cannot execute code or request additional data.

Each hypothesis must specify a natural-language claim, all variables
involved, the claimed direction, the claimed absolute change in
outcome probability, and an optional condition or population.
Do not assign benchmark archetype names. Avoid duplicate or
paraphrased claims.

Return:
[
  {
    "text": "<one-sentence testable hypothesis>",
    "variables": ["<variable>", "..."],
    "direction": "<positive, negative, or mixed>",
    "magnitude": <absolute probability change between 0 and 1>,
    "condition": "<condition or population, or null>"
  }
]
\end{verbatim}
\end{promptbox}

\begin{promptbox}{XGBoost+LLM}
\begin{verbatim}
SYSTEM:
You are a data scientist converting predictive evidence into
testable hypotheses. Return JSON only.

USER:
Domain: {domain}
Dataset description: {scenario}
Outcome: {outcome}

Schema:
{schema}

Variable distributions:
{distribution_summaries}

Random sample of 1,000 rows:
{sample_rows}

Features ranked by XGBoost gain importance:
{ranked_features}

Using the observed examples, distribution summaries, and feature
ranking, generate {k} distinct, testable hypotheses about variables
that may affect the outcome. Feature importance is predictive rather
than causal evidence. You cannot execute code or request additional
data.

Each hypothesis must specify a natural-language claim, all variables
involved, the claimed direction, the claimed absolute change in
outcome probability, and an optional condition or population.
Do not assign benchmark archetype names. Avoid duplicate or
paraphrased claims.

Return:
[
  {
    "text": "<one-sentence testable hypothesis>",
    "variables": ["<variable>", "..."],
    "direction": "<positive, negative, or mixed>",
    "magnitude": <absolute probability change between 0 and 1>,
    "condition": "<condition or population, or null>"
  }
]
\end{verbatim}
\end{promptbox}

\end{document}